\documentclass{ecai} 

\usepackage{latexsym}
\usepackage{amssymb}
\usepackage{amsmath}
\usepackage{amsthm}
\usepackage{booktabs}
\usepackage{graphicx}
\usepackage{color}

\usepackage{subfig}
\usepackage{colortbl}
\usepackage{algorithm}%
\usepackage{algorithmicx}%
\usepackage{algpseudocode}%
\usepackage{multirow}
\usepackage[normalem]{ulem}
\usepackage{xcolor}
\usepackage[utf8]{inputenc}
\usepackage{tcolorbox}
\usepackage{listings}
\usepackage{amsmath}
\usepackage{enumitem}
\definecolor{mykeywordgreen}{RGB}{0,128,0} 
\definecolor{mygreen}{RGB}{0,128,0} 

\def\BibTeX{{\rm B\kern-.05em{\sc i\kern-.025em b}\kern-.08em
    T\kern-.1667em\lower.7ex\hbox{E}\kern-.125emX}}

\newcommand{\ours}{\ensuremath{{\sf HalMit}}}
\newif\ifmark
\newif\ifhidenote
\marktrue
\hidenotefalse

\usepackage[normalem]{ulem}
\usepackage{xcolor}
\ifmark
  \newcommand{\rev}[1]{\textcolor{blue}{{#1}}}

    \newcommand{\del}[1]{\sout{#1}}
  \ifhidenote
     \newcommand{\note}[1]{}
  \else
  \newcommand{\note}[1]{\normalem\emph{\textbf{\uline{#1}}}}
  \fi
\else
  \newcommand{\rev}[1]{{#1}}
  \newcommand{\del}[1]{}
  \newcommand{\note}[1]{}
\fi

\begin{document}


\begin{frontmatter}


\paperid{3227} 


\title{Towards Mitigation of Hallucination for LLM-empowered Agents: Progressive Generalization Bound Exploration and Watchdog Monitor}


\author[A,B]{\fnms{Siyuan}~\snm{Liu}}
\author[C]{\fnms{Wenjing}~\snm{Liu}}
\author[B,D]{\fnms{Zhiwei}~\snm{Xu}\thanks{Corresponding Author. Email: xuzhiwei2001@ict.ac.cn.}} 
\author[E]{\fnms{Xin}~\snm{Wang}}
\author[F]{\fnms{Bo}~\snm{Chen}}
\author[A]{\fnms{Tao}~\snm{Li}\thanks{Corresponding Author. Email: litao@nankai.edu.cn.}} 

\address[A]{College of Computer Science, Nankai University}
\address[B]{Haihe Lab of ITAI}
\address[C]{College of intelligent Science and Technology, Inner Mongolia University of Technology}
\address[D]{Institute of Computing Technology, Chinese Academy of Sciences}
\address[E]{Department of Electrical and Computer Engineering, Stony Brook University}
\address[F]{Department of Computer Science, Michigan Technological University}


\begin{abstract}
Empowered by large language models (LLMs), intelligent agents have become a popular paradigm for interacting with open environments to facilitate AI deployment. However, hallucinations generated by LLMs—where outputs are inconsistent with facts—pose a significant challenge, undermining the credibility of intelligent agents. Only if hallucinations can be mitigated, the intelligent agents can be used in real-world without any catastrophic risk. Therefore, effective detection and mitigation of hallucinations are crucial to ensure the dependability of agents. Unfortunately, the related approaches either depend on white-box access to LLMs or fail to accurately identify hallucinations. To address the challenge posed by hallucinations of intelligent agents, we present HalMit, a novel black-box watchdog framework that models the generalization bound of LLM-empowered agents and thus detect hallucinations without requiring internal knowledge of the LLM’s architecture. Specifically, a probabilistic fractal sampling technique is proposed to generate a sufficient number of queries to trigger the incredible responses in parallel, efficiently identifying the generalization bound of the target agent. Experimental evaluations demonstrate that HalMit significantly outperforms existing approaches in hallucination monitoring. Its black-box nature and superior performance make HalMit a promising solution for enhancing the dependability of LLM-powered systems.
\end{abstract}

\end{frontmatter}


\section{Introduction}
\label{sec:intro}

With the rapid proliferation of artificial intelligence in contemporary life, agents empowered by large language models (LLMs) have emerged as pioneers of this technology transformation~\cite{yang2024harnessing}.
However, in conjunction with their widespread deployment, the phenomenon of hallucination has become a major concern in LLM and their agents~\cite{huang2023surveyhallucinationlargelanguage}. LLM hallucination refers to instances in which LLM generated content is inconsistent, unfaithful, contradictory, or unverifiable against established real-world knowledge \cite{huang2023surveyhallucinationlargelanguage}, although it may be presented in a convincing and confident tone. This issue has been recognized by many academic studies and technical public reports as one of the primary ethical and safety risks associated with LLM agents, along with issues such as bias and toxic content. The hallucination phenomenon becomes thorny when considering the black-box nature of LLMs, and severely undermines the credibility of LLM agents, especially in truth-sensitive fields such as law ~\cite{greco2023bringing}, medicine~\cite{freyer2024future}, finance~\cite{zhao2024revolutionizing}, and education~\cite{kasneci2023chatgpt}, where it can have catastrophic cognitive consequences.

Mitigating hallucinations is critical to improving the dependability of LLM agents in real-world applications~\cite{huang2023surveyhallucinationlargelanguage}. Hallucinations typically arise when the generated content significantly exceeds the generalization bounds of the agent~\cite{yang2024exploring}. 
If a generated response lies outside the bound, it is highly likely that this response is hallucinated~\cite{zhang2024knowledge}. Therefore, identifying the generalization bounds is of critical importance in mitigating hallucinations in LLM agents.
Although recent efforts have been focused on computing non-vacuous generalization bounds for deep learning models, these bounds tend to become vacuous at the scale of billion-parameter models~\cite{lotfi2024unlocking,zhang2024knowledge}. In addition, such theoretical bounds are often derived from restrictive statistical assumptions that limit their applicability to models with low generalization capacity. Given the vastness of the semantic space, the tightness of the existing generalization bounds remains difficult to establish \cite{lotfi2023non}.

Existing approaches attempt to identify hallucinated responses by analyzing the internal state of the model \cite{ji2024llm, han2024semantic, zhu2024pollmgraph, he-etal-2024-llm}. This requires full transparency and access to the internal state of the model, known as ``white-box access'', and cannot work for large-scale commercial models, which are usually close-sourced.
The other approaches~\cite{andriopoulos2023augmenting, zhang2023user, wei2024measuring} are mainly based on cross-checking of the LLM output against external databases. Recent work for modeling generalization bound computes non-vacuous generalization bounds for deep learning models, but these bounds are vacuous for large models at the billion-parameter scale ~\cite{hou2024probabilistic, quevedo2024detecting} that involve directly asking the LLM to produce its confidence scores in the truthfulness of the statements (referred to as ``model confidence'' in this paper). Although these black-box/gray-box approaches allow hallucination monitoring through output text or associated confidence scores, they are degraded by limited knowledge of LLMs and often poorly calibrated confidence estimates. Ultimately, these score-based approaches may be incorrect, reducing the effectiveness and accuracy of the monitoring. Therefore, there is a strong demand for developing new techniques that can mitigate hallucinations while avoiding the aforementioned limitations.

\begin{figure}[ht]
\centering
\includegraphics[width=0.63\linewidth]{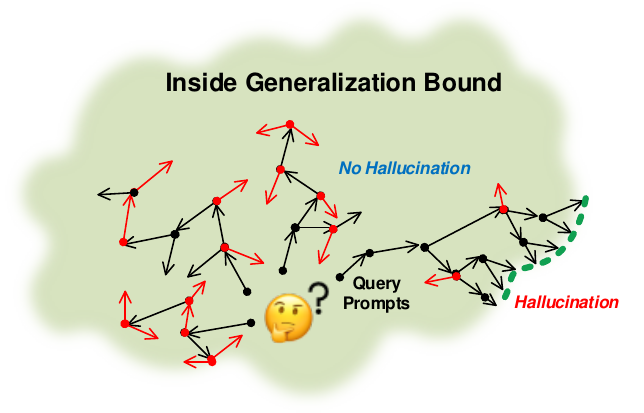}

\caption{\label{introfig} The complexity of generalization bound modeling.}
\vspace{0.4 cm}
\end{figure}

Due to the complexity of the semantic space, deriving a universal generalization bound across all domains is extremely challenging~\cite{yang2024exploring}. However, we observe that this bound may be identified much more easily within the domain of each specific agent (Section 2). 
Building on this insight, we propose \textit{$\ours$}, a fine-grained approach for modeling per-agent generalization bounds to monitor hallucinations that fall outside these boundaries. A major challenge in designing \textit{$\ours$} lies in the need to efficiently identify and model the complex generalization bound of the target agent in a specific domain. As shown in Figure \ref{introfig}, the generalization bound is difficult to pinpoint. The bound exploration process may deviate from the true boundary and become trapped in local loops (red arrows), considering that the bounded space itself is vast. 

To accelerate the generalization bound exploration, we propose a probabilistic fractal sampling method to generate a sufficient number of parallel queries so that they can efficiently cover the generalization bound of the targeted agent. 
Each identified boundary point is stored in a vector database,  where the point is represented by its query–response pair along with associated context. 
During hallucination monitoring, $\ours$ compares the input query with those retrieved from the vector database. If the query closely resembles the retrieved records in the vector base, it is considered near the boundary, and the response corresponding to the input query is flagged as a potential hallucination. 
Our major contributions are summarized as follows:
\begin{enumerate}[label=\arabic*)]
\vspace{-0.1 cm}
    \item Through a preliminary study on agent hallucinations, we confirm that hallucinated responses correspond to the agent’s generalization bound and that it is possible to model this bound within specific application domains. 
    \item We propose a novel probabilistic fractal exploration scheme to enable our MAS system to incrementally probe the generalization boundary. In this process, deep reinforcement learning guides the multi-agent exploration by adjusting the probabilities of fractal transformations based on three common semantic patterns. These probabilities are continuously updated to efficiently and effectively model the generalization bound within a specific domain.  
    \item A unique hallucination mitigation technology is provided based on the generalization boundary to enable more dependable monitoring and persistently detecting potential hallucinations.
    \item We have conducted extensive experimental evaluations and the results are encouraging, demonstrating a significant improvement in hallucination monitoring effectiveness over baseline solutions. A unique hallucination mitigation technology is provided to enable a more dependable monitoring and monitoring of potential hallucinations.
\end{enumerate}
\vspace{-0.1 cm}

These contributions address vital challenges in mitigating hallucination of LLM-empowered agents, paving the way for more trustworthy intelligent systems in real-world applications. To the best of our knowledge, this is the first hallucination monitoring approach that operates without access to internal model knowledge or reliance on cross-verification algorithms. This design enables real-time and persistent hallucination mitigation in deployed intelligent systems. The source code for $\ours$ will be available on GitHub after the paper is accepted.
\section{Motivation}

In this section, we present preliminary studies that investigate how hallucinations manifest in LLM-empowered agents across various domains. 
\begin{itemize}
    \item \textbf{PS1}: We investigate whether the statistic characteristics of hallucinations vary across different domains.
    \item  \textbf{PS2}: We examine the statistic characteristics within each domain to identify potential patterns or regularities in LLM hallucinations.
    \item  \textbf{PS3}: We assess whether certain  existing characteristics can be directly leveraged to monitor hallucinations.
\end{itemize}
\vspace{-0.6 cm}
 \begin{figure}[ht]
    \centering
    \subfloat[Health]{
        \label{fig:subfig1}
        \includegraphics[width=0.4\linewidth]{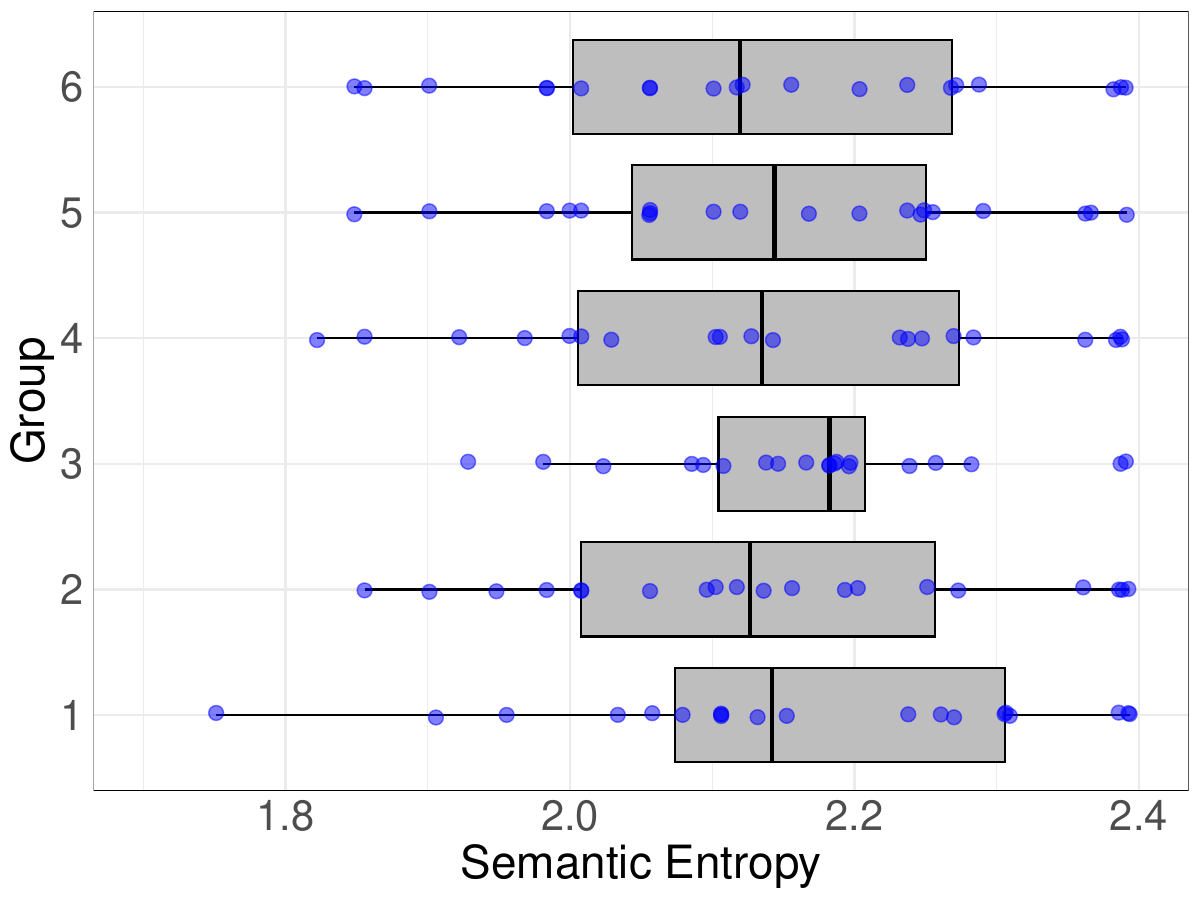}
    }
    \subfloat[Nutrition]{
        \label{fig:subfig2}
        \includegraphics[width=0.4\linewidth]{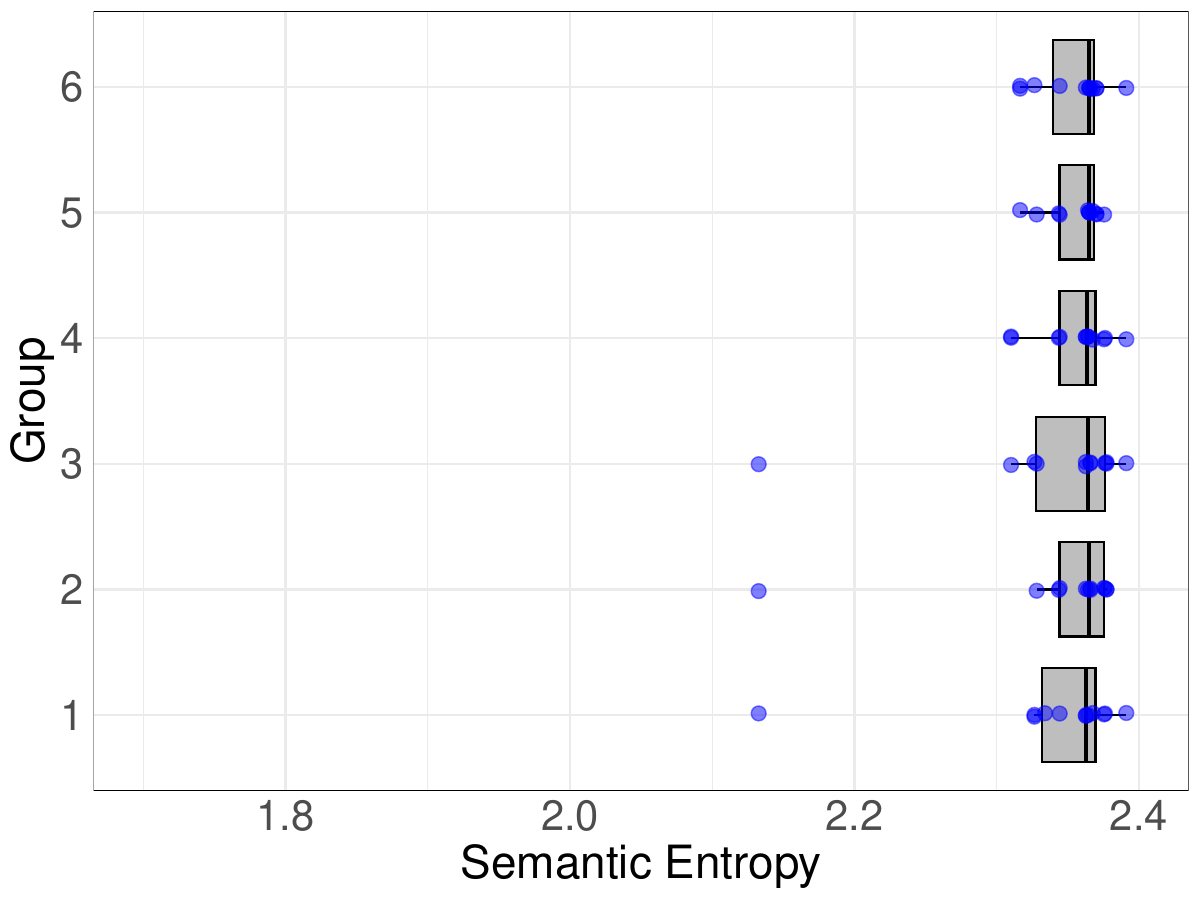}
    }

    \vspace{-0.3cm} 
    
    \subfloat[Sociology]{
        \label{fig:subfig3}
        \includegraphics[width=0.4\linewidth]{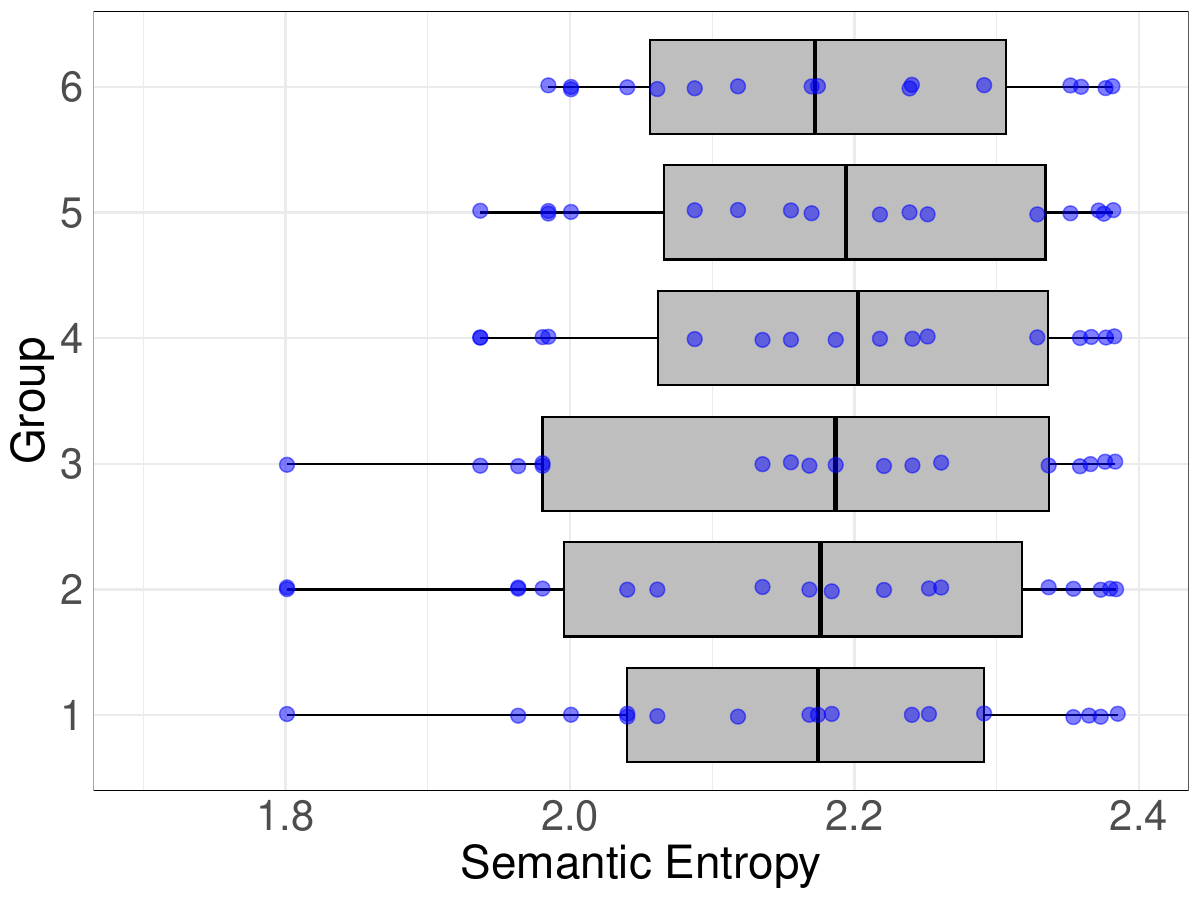}
    }
    \subfloat[Law]{
        \label{fig:points:3}
        \includegraphics[width=0.4\linewidth]{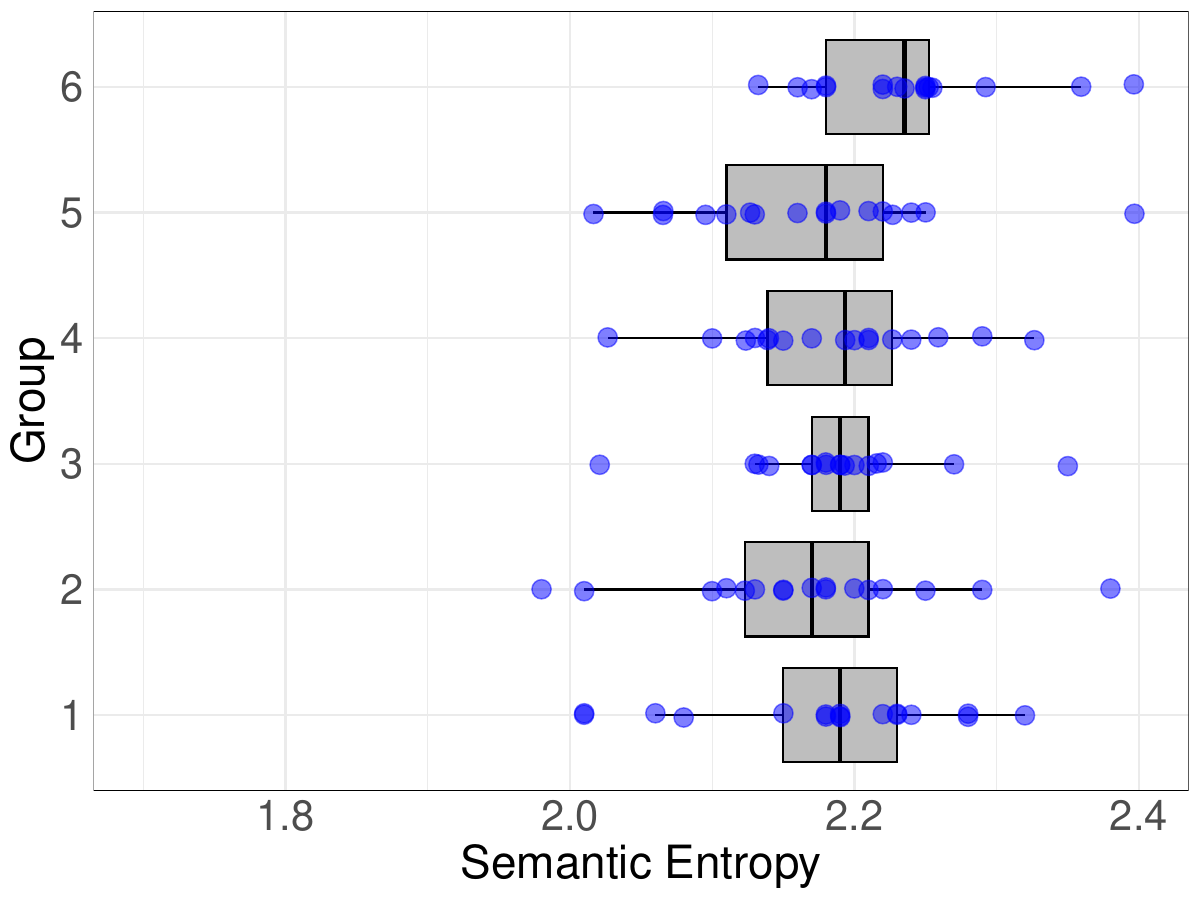}
    }
    \vspace{-0.3cm} 
    \subfloat[Fiction]{
        \label{fig:points:2}
        \includegraphics[width=0.4\linewidth]{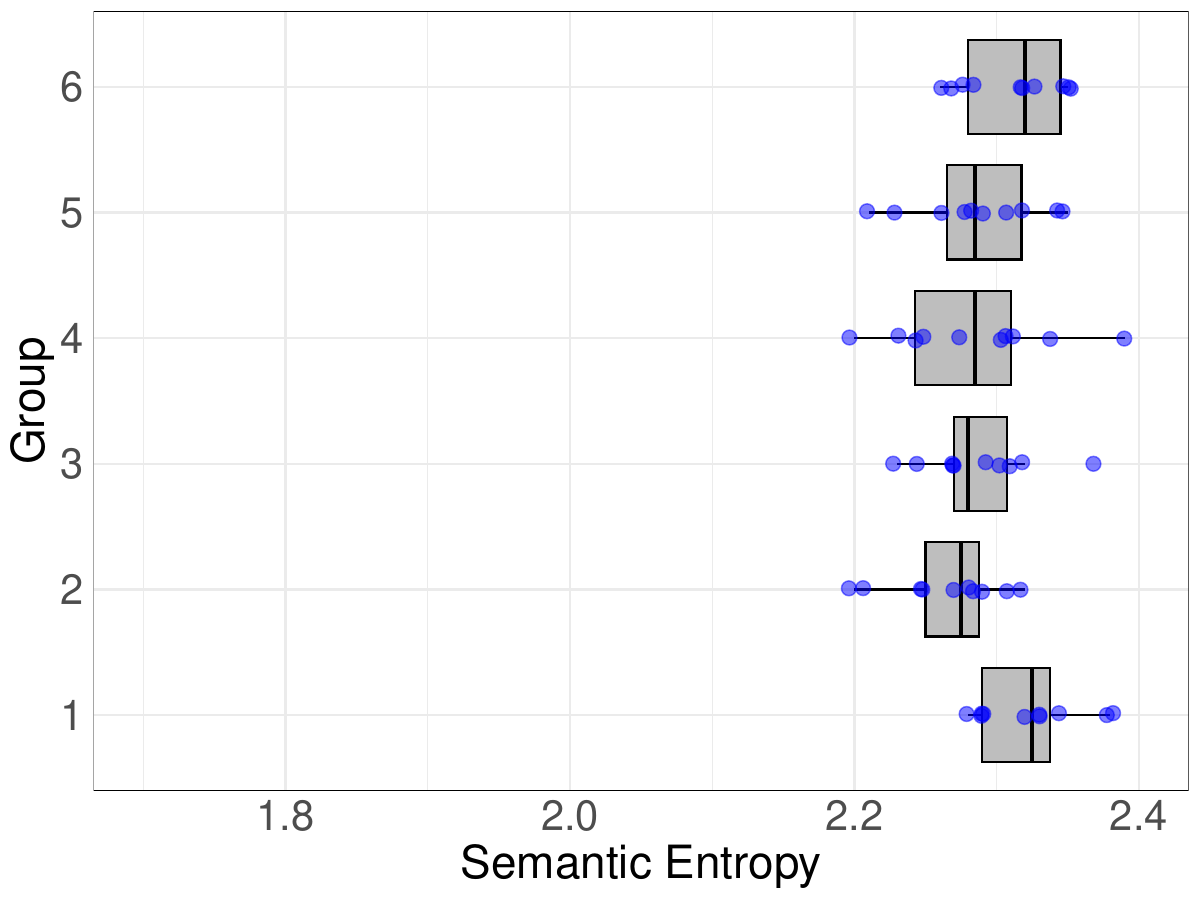}
    }
    \subfloat[Paranormal]{
        \label{fig:subfig6}
        \includegraphics[width=0.4\linewidth]{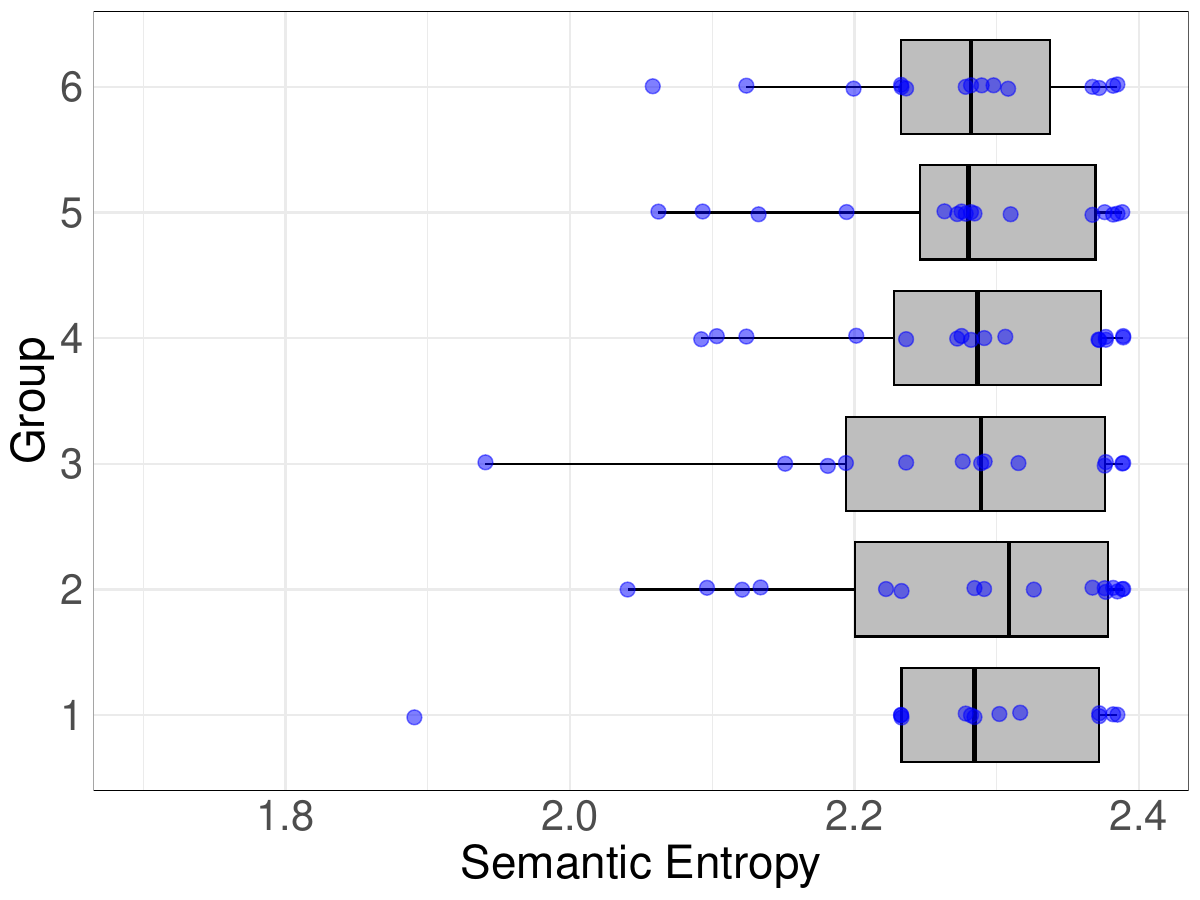}
    }
    
    \caption{The semantic entropy values for six popular domains.}
    \label{fig:points}
    \vspace{0.3cm} 
\end{figure}

\noindent\textbf{Experimental Settings:} Without loss of generality, we study the response quality of agents powered by Llama3.1-8B across six popular domains, including health, nutrition, sociology, law, fiction, and paranormal. All query-answer pairs and domains used in the study are randomly sampled from a real-world hallucination dataset, TruthfulQA ~\cite{lin2022truthfulqa},  which includes both truthful and hallucinated responses. 
Semantic entropy~\cite{farquhar2024detecting} is one 
metric that has been extensively used to assess the uncertainty level of agent responses (see Appendix B), with higher entropy value typically indicating greater uncertainty and a higher likelihood of hallucinations. Unlike general confidence scores of LLM's response, semantic entropy is specifically designed to characterize the hallucinations of LLMs. Based on semantic entropy, we evaluate response quality across the six domains by measuring semantic entropy, with the results presented in Figure~\ref{fig:points}.

\begin{itemize}
\vspace{-0.2 cm}
    \item \textbf{PS1}: \textbf{Significant Variations across Domains.} As shown in Figure \ref{fig:points}, the semantic entropy values of agent responses vary substantially across application domains, with noticeable differences in both  medians and variances. This indicates that the statistical characteristics of hallucinations differ significantly by domains, suggesting that no universal generalization bound can be established across all domains.

    \item  \textbf{PS2}:  \textbf{Statistic Stability within The Same Domain.}
Further analysis of the boxplots reveals that within each individual domain, the semantic entropy values, while subject to some fluctuation across testing groups, tend to follow a consistent distribution.  This internal stability suggests that hallucination patterns are coherent within each domain, making it feasible to identify a domain-specific generalization bound. 

\item  \textbf{PS3}: \textbf{Limitations of a Threshold Based on the Existing Metric.}
As observed in the boxplots for the Health, Nutrition, Law, Fiction, and Paranormal domains, outliers extending beyond the whiskers in the boxplots are present. This suggests that while semantic entropy has been shown to support hallucination detection~\cite{kuhn2023semantic, farquhar2024detecting, han2024semantic}, relying solely on a fixed threshold is insufficient. The presence of high-entropy yet potentially non-hallucinatory responses (and vice versa) highlights the need for more nuanced detection methods.
\vspace{-0.2 cm}
\end{itemize}

\vspace{-0.1 cm}
\noindent\textbf{Summary of findings: }
Based on the above observations, our findings reveal that: (1) hallucination patterns vary across application domains but tend to exhibit consistent statistical behavior within the same domain; and (2) hallucinations cannot be effectively detected using a simple threshold on any single existing metric, even one as significant as semantic entropy. 
These insights suggest that the generalization bound distinguishing hallucinated from non-hallucinated responses is more clearly defined when tailored to a specific agent within a specific domain, rather than derived from a collective set of agents or domains. Consequently, accurate hallucination monitoring requires the identification of these generalization bounds in a fine-grained, domain-aware manner.

\textbf{Motivated by these observations, we propose \textit{HalMit}, a hallucination mitigation paradigm for LLM-empowered agents.}

\section{Methodology}
\label{sec:method}
To persistently mitigate hallucinations, $\ours$ functions as a ``watchdog'' framework for each target agent to monitor hallucinations. Before being used to monitor hallucinations, $\ours$ first models the agent’s  generalization bound based on a proposed multi-agent exploration system.
This allows hallucinations to be identified and mitigated based on their deviation from the learned generalization boundary.

\subsection{Generalization Exploration Bound with a MAS}
Given the black-box nature of LLM-empowered agents, exploring their generalization bounds and identifying hallucinations are critical and challenging~\cite{zhang2021understanding}. To address this, we introduce a multi-agent bound exploration method that integrates probabilistic fractal sampling into a multi-agent system (MAS) for parallel query generation. To improve the efficiency and relevance of the queries, we propose a reinforcement learning-based scheme to dynamically adjust the fractal probabilities.

\begin{figure}[ht]
\centering
\includegraphics[width=0.95\linewidth]{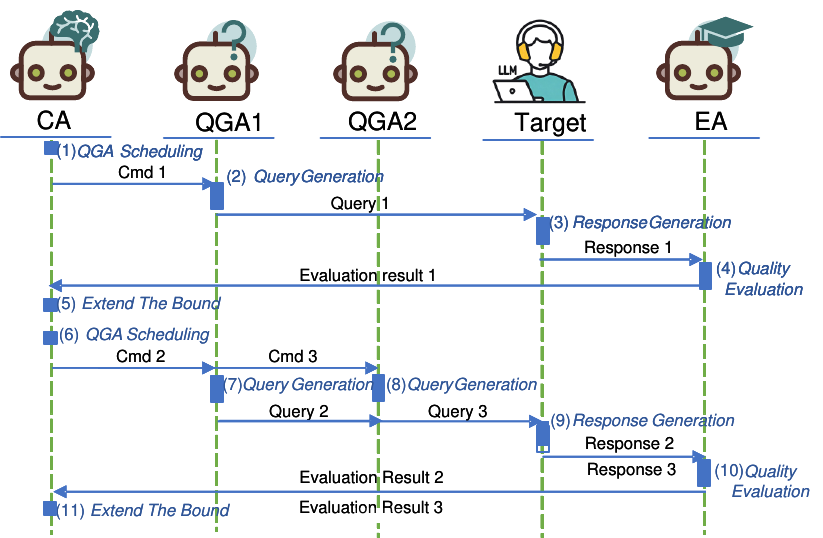}
\caption{\label{fig:framwork} Multi-agent collaboration in HalMit.}
\vspace{0.4 cm}
\end{figure}

As illustrated in Figure \ref{fig:framwork}, the proposed MAS consists of three specialized agent types: core agent (CA), query generation agent (QGA), and evaluation agent (EA).
Among them, the CA coordinates  interactions between QGAs and the target LLM-powered  agent. Meanwhile, the EA, guided by HalluBench criteria~\cite{zhao2023beyond}, evaluates  the quality of the response from the target and provides essential feedback to refine the query generation process.
\label{sec:framework-overview}

\subsubsection{Probabilistic Fractal-based Query Generation}
\label{sec:Prompt_Generation}

Leveraging the self-similarity feature of the fractal in natural language, we propose a novel probabilistic fractal-based query generation method  for use in QGAs. This method iteratively constructs increasingly complex query structures that progressively approach the generalization bound of the target agent. Unlike conventional fractal systems that apply all affine transformations in each iteration step, our method proposes an Iterated Function System with Probabilities (IFSP) (See appendix A) 
that enough queries can be generated quickly to cover the generalization bound of the target agent $\tau$. More specifically, according to semantic theory~\cite{riemer2002remetonymizing}, three semantic extension patterns, \textit{induction}, \textit{deduction}, and \textit{analogy}, are used as fractal affine transformations to extend specific queries and navigate the semantic space.
The execution probability of each transformation is dynamically adjusted based on the IFSP system:
\begin{equation}
   \mathcal{F}=\{FTi:\mathcal{P}_{t-1}^{\tau}\overset{p_i}{\rightarrow} \mathcal{P}_{t}^{\tau},\quad \sum_{i=1}^3p_i=1\}, \\
\end{equation}
where ${P}_{t}^{\tau}$ are queries used in round $t$, functions $FT1, FT2, FT3$ correspond to three fractal affine transformations, $p_i$ is the execution probability for $FTi$. These three types of fractal affine transformations are introduced as follows:

\begin{itemize}
\vspace{-0.4 cm}
    \item \textbf{FT1: Semantic Deduction.} This transformation generates more specific queries by deriving them from general rules or concepts presented in the previous iteration. For example, given a query, \textit{“Did humans really land on the moon in 1969?”}, a deductive transformation would produce a more focused follow-up such as\\\textit{“What were the technological advancements that enabled humans to land on the moon in 1969?”}.

    \item \textbf{FT2: Semantic Analog.} This transformation broadens the scope of the original query by leveraging  semantic associations such as synonyms, antonyms, or functional analogies. In this way, a new query, \textit{“What historical events in space exploration paralleled the significance of the moon landing in 1969?”}, can be generated. This helps the system probe parallel narratives and conceptual similarities in the semantic space.

    \item \textbf{FT3: Semantic Induction.} This transformation generates broader, more abstract queries by generalizing from specific instances and inferring underlying linguistic or conceptual patterns. As another example, \textit{“How can we assess the reliability of commonly accepted events in the history of space exploration?”}, is obtained by following FT3. This supports exploration of overarching themes and epistemological questions within the domain. 

\end{itemize}
\vspace{-0.1 cm}



These three affine transformations are applied in each iteration of the exploration process. In this way, new queries are generated through the iterative process that starts from basic concepts or principles.
The core agent can order multiple query generation agents to generate queries,  so the iteration can be realized through parallel processes, to significantly increase the speed in identifying the generalization bound. 
This iteration process includes four steps:
\vspace{-0.2 cm}
\begin{enumerate}[label=\arabic*)]
    \item To cover a broader semantic space with in the bound, the CA randomly initializes multiple queries in a domain and sends each query to the target agent $\tau$ as initial questions.  
    \item The target agent responds to queries with responses that may contain hallucinations. Therefore, each QA pair is sent to an EA. After receiving the QA pair, the EA assesses whether the response in the received QA pair contains a hallucination, and sends a report back to the CA, including the QA pair and the corresponding evaluation results. 
    \item Depending on the evaluation result, QGA will perform query generation in two ways. In case a hallucination is reported, the CA embeds the QA pair and the context information into a vector database as a point of the generalization bound of agent $\tau$ (detailed in Section~\ref{sec:detection}). To expand the exploration range and cover more of the generalization bound, the CA schedules the QGAs to generate new queries through the fractal affine transformations, FT1 and FT2, where the probability of each is determined through reinforcement learning (details in Section \ref{sec:Reinforced}). Otherwise, in case no hallucination is reported, new queries will be randomly generated, and sent back to the CA. 
    \item The new queries generated by the QGAs are sent back to the target agent. In this way, a new round of fractal exploration with a pipeline is scheduled by CA, where multiple new iterations are generated in parallel in response to each exploration path. More EAs are scheduled to assess all QA pairs and the evaluation reports will be sent to the CA. The bound search speed can be exponentially increased in this way.
\end{enumerate}
\vspace{-0.2 cm}
During this iterative process, the ratio of the hallucinations among all QA pairs, $\gamma$, is incrementally updated by the core agent. Once $\gamma$ becomes larger than an empirical threshold $\epsilon$, it indicates that the generalization bound of agent $\tau$ can be identified by the vector database, and this iterative process of $\mathcal{F}$ ends. These parameters are evaluated in ablation studies in Section \ref{sec:ablation}. 

\vspace{-0.2 cm}
\subsubsection{Reinforced Determination of Fractal Probabilities}
\begin{figure}[ht]
\centering
\includegraphics[width=0.95\linewidth]{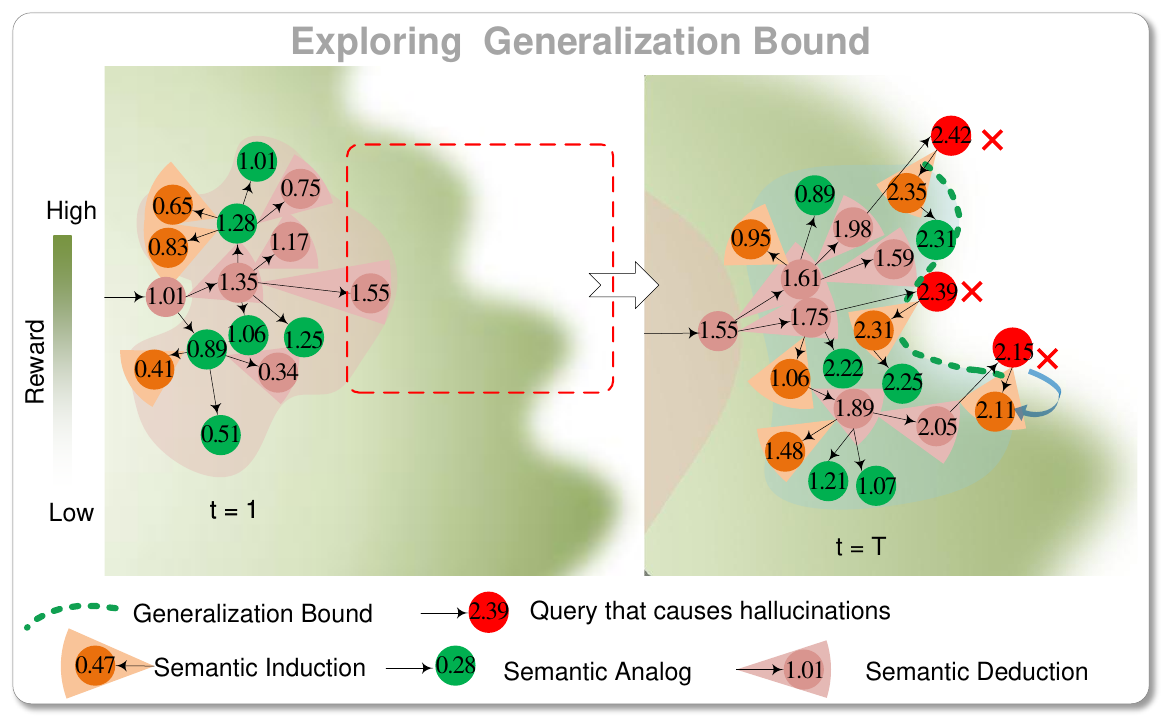}
\caption{\label{fig:rl}Process of exploring generalization bound. The numbers in the circles are the corresponding semantic entropies.}
\vspace{0.4 cm}
\end{figure}

\vspace{-0.1 cm}
\label{sec:Reinforced}
To further increase the efficiency in identifying the generalization bound, we use deep reinforcement learning \cite{UDEKWE2024123055} to determine the probability of each transformation function in IFSP $\mathcal{F}$ to go for in the next step. This probability is adjusted in each iteration so that the exploration process can more efficiently converge towards the bound. Deep reinforcement learning is a framework in which a policy network is trained to maximize long-term objectives.
In our design, the policy network is trained to efficiently select the appropriate probabilities of fractal affine transformations in each iteration, driving the fast convergence to the bound as its long-term objective, as shown in Figure \ref{fig:rl}. This design significantly accelerates the exploration process.

\noindent\textbf{Training Data Prepared for Policy Network: }
\label{Generation Train Data}

To evaluate the effectiveness of a fractal affine transformation in state $s_i$, we repetitively send the corresponding query, $\mathcal{P}^\tau_i$, to the target agent $K$ times, collect $K$ responses, $\{a^\tau_i\}_K$, and according to \cite{farquhar2024detecting}, calculate the semantic entropy for the target agent in state $s_i$, $H^\tau_i$. 
More specifically, every item in $\{a^\tau_i\}_K$, $a^\tau_{i}(k)$, is included in the token sequence group for calculating $H^\tau_i$. Define a function $\operatorname{sig}(a^\tau_i(k))$, which is equal to 0 if the answer of the target agent contains hallucinations; otherwise, $\operatorname{sig}(a^\tau_i(k))$ is equal to 1. Hence, the reward for the triple $\{\mathcal{P}^\tau_i,{\{a^\tau_i\}}_K,  H^\tau_{i-1}\}$ can be given as:
\begin{equation}\label{reward}
R_i^\tau\left(\mathcal{P}_i^\tau,\left\{a_i^\tau\right\}_K, H_{i-1}^\tau, H_i^\tau\right) \\
= \begin{cases}\Delta H_i^\tau  & \text { if } \prod_{k=1}^K \operatorname{sig}\left(a_i^\tau(k)\right)^\tau \neq 0 \\ \left|\frac{1}{R_{i-1}^\tau}\right| & \text { if } \prod_{k=1}^K \operatorname{sig}\left(a_i^\tau(k)\right)^\tau=0\end{cases}
\end{equation}

We generate queries based on three fractal affine transformations, collect their queries, answers, and semantic entropy, and calculate their rewards. 
Considering that each of these rewards represents the degree to which the enhancement in the exploration of generalization bounds is achieved through the corresponding affine transformation, we configure the probability for an affine transformation $j$ ($j\in \{1,2,3\}$) in our IFSP as:
\begin{equation}\label{p}
    p_j = R_j^\tau /\sum_{k=1}^3{R_k^\tau}.
\end{equation}
Each $p_j$, along with the corresponding input query, response, and semantic entropy, is included in the training database of the policy network, which consists of a quadruple, $\{\mathcal{P}^\tau_j,{\{a^\tau_j\}}_K,  H^\tau_{j-1}, p_j\}$. Through the iteration process of fractal sampling on the target agent, more quadruples can be collected until the scale of this dataset is sufficient for the policy network training.

\noindent\textbf{Policy Network Training: }
A popular design of the policy network multilayer perceptron (MLP)~\cite{UDEKWE2024123055} is introduced to capture the nonlinear representation of state features and predict the probability distribution of three affine transformations $\{p_j\}$ in IFSP $\mathcal{F}$. States in the generalization space of the target agent are defined as:
\begin{equation}
  s_i^\tau=\lfloor{ \frac{\Delta\text{Sim}(\mathcal{P}^\tau_0, \mathcal{P}^\tau_i)\times e^{H^\tau_i}}{\omega}}\rfloor,
\end{equation}
where $\mathcal{P}^\tau_0$ is the initial query used to explore the generalization bound. Its semantic similarity to $\mathcal{P}^\tau_i$ is involved as a scaling parameter and $\omega$ is a normalization parameter.  

For each state $s_i^\tau$, the action-value function $Q()$ for affine transformations $f_i^\tau$ at the time step $i$ is  
$  Q(s_i^\tau, f_i^\tau) = \mathbb{E}\left[R_i^\tau \mid s_i^\tau, f_i\right] $, 
where $R_i^\tau$ is the reward defined in Formula ~(\ref{reward}). 
The objective $Q\left(s_i^\tau, f_i; \theta_i\right)$ is to train the policy network, parameterized by $\theta$ denoted. The network is optimized to align with the optimal action-value $Q^*(s_i^\tau, f_i)$, which corresponds to the highest possible reward.  To achieve this, a loss function is defined using the L2 norm:  
\begin{equation}  
  \mathcal{L}\left(\theta\right) = \sum_{i=1}^{N}\left\|Q^*(s_i^\tau, f_i) - Q\left(s_i^\tau, f_i; \theta_i\right)\right\|,
\end{equation}  
where $N$ is the mini-batch of training samples. 
Finally, the optimal parameters $\theta^*$ of the policy network can be obtained by:  
\begin{equation}  
  \theta^* = \arg \min_{\theta_i} \sum_{i=1}^{N} \mathcal{L}\left(Q^*(s_i^\tau, f_i), Q\left(s_i^\tau, f_i ; \theta_i\right)\right).  
\end{equation}

The convergence study
of the exploration is provided in the experimental section (Section \ref{sec:Convergence}).

\subsection{Hallucination Monitoring}
\label{sec:detection}

In this section, we describe how the generalization bound can be leveraged to monitor if there exist hallucinations in the response from a target agent.
This is achieved by comparing the response with the information retrieved from the vector database that represents the generalization bound of the target agent. As the bound often has an irregular shape (as illustrated in Figure \ref{introfig}), hallucination monitoring can be very difficult.

\begin{figure}[ht]
\centering
\includegraphics[width=0.5\linewidth]{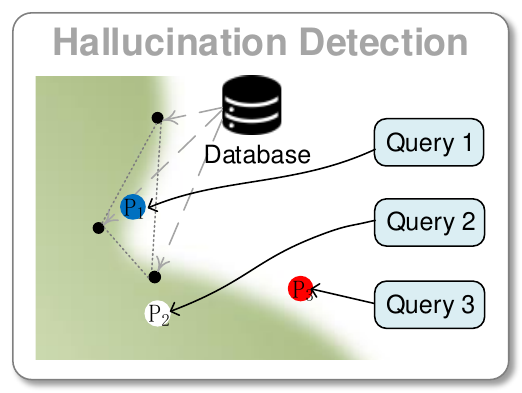}
\caption{\label{fig:dect} Hallucination monitoring}
\vspace{0.3 cm}
\end{figure}

Hallucinated responses tend to diverge significantly, making it difficult to achieve a meaningful comparison with the bound. Instead, we choose the use of input query to compare against the bound.
To start hallucination monitoring, the input query $P_q^\tau$ is compared with all related items in the vector database built in Section~\ref{sec:method}. This is achieved by evaluating the cosine similarity of the query vector $Q_v $ and each related vector $v_i$ denoted by $S_i$. Specifically, we consider three cases:

\begin{itemize}
    \item When there are more than three similar items in the database that exceed a threshold $\epsilon$, we calculate the centroid of three most similar items $P^\tau_{v_1}, P^\tau_{v_2}, P^\tau_{v_3}$:
    \begin{equation}\label{fum19}
        C = \frac{\sum_{i=1}^3 S_i \cdot v_i}{\sum_{i=1}^3 S_i},
    \end{equation}
where $S_i$ is the similarity score of the $i$-th vector. 
Next, we calculate the cosine similarity $S_C$ between the query vector $Q_v$ and the normalized centroid $C$. If $S_C > \epsilon$, the input query is considered to be beyond the generalization bound of the target agent (corresponding to the blue point in Figure \ref{fig:dect}).
\item Otherwise, we compare the semantic entropy of the query $H_{Q_v}^{\tau}$ with the semantic entropy of the most similar vector $H^{\tau}_{vi}$ in the vector database. If $H_{Q_v}^{\tau}$ is larger, the input query is likely to be outside the generalization bound and may cause a hallucination (corresponding to the red point in Figure \ref{fig:dect}e). 
\item If $H_{Q_v}^{\tau} < H^{\tau}_{vi}$, the input query is within the generalization bound, and will obtain a rational response (corresponding to the white point in Figure \ref{fig:dect}).
\end{itemize}
The details of the monitoring process are given in Algorithm \ref{alg1}.

\begin{algorithm}[htbp]
\footnotesize
\caption{\label{alg1}Hallucination monitoring algorithm}
\begin{algorithmic}[1]

\State \textbf{Input:} Input query $P_q^\tau$, vector database $\mathcal{V}^\tau$, similarity threshold $\epsilon$ of hallucination monitoring;
\State \textbf{Output:} Hallucination status of $P_q^\tau$;

\For{each vector $v_i \in \mathcal{V}^\tau$} 
    \State Normalize $v_i$: $v_i \gets \frac{v_i}{\|v_i\|}$;
\EndFor

\State Represent $P_q^\tau$ with a vector: $Q_v^\tau \gets \text{Embedding}(P_q^\tau)$;
\State Normalize $Q_v^\tau$: $Q_v^\tau \gets \frac{Q_v^\tau}{\|Q_v^\tau\|}$;

\State Initialize an empty list of $results$;
\For{each vector $v_i \in \mathcal{V}^\tau$}
    \State Compute cosine similarity $S_i$;
    \State Append $(v_i, S_i)$ to $results$;
    \State Sort $results$ by similarity score $S_i$ in descending order;
\EndFor

\If{$results[3] > \epsilon$}


\State Compute the centroid by Formula (\ref{fum19});
\State Normalize the centroid: $C \gets \frac{C}{\|C\|}$;

\EndIf
\State Compute the similarity $S_C$ between $Q_v^\tau$ and the centroid $C$;

\If{$S_C \geq \epsilon$}
    \State Report $P_q^\tau$ may cause a hallucination;
\Else
    \State Compute the semantic entropy $H(Q_v^\tau)$ of the query;
    \If{$H(Q_v^\tau) > max(H(v_i^\tau))$}
        \State Report $P_q^\tau$ may cause a hallucination;
    \Else
        \State Return the response of the agent for $P_q^\tau$; 
    \EndIf
\EndIf
\end{algorithmic}
\end{algorithm}

\vspace{-0.4 cm}
\section{Experimental Evaluation}
\label{exp}

\subsection{Datasets}
To evaluate the performance of $\ours$, two popular public Query-Answer (QA) datasets, MedQuAD \cite{ben2019question} and SQuAD~\cite{rajpurkar-etal-2016-squad}, are used.
\vspace{-0.1 cm}
\begin{itemize}
\item MedQuAD is a collection of question-answer pairs meticulously curated from 12 trusted National Institute of Health (NIH) websites and covers various medical topics including diseases, medications, and diagnostic tests.
\item SQuAD consists of questions posed by crowd-workers on a set of Wikipedia articles, where the answer to every question is a segment of text or span, and each QA pair is paired with a title.
\end{itemize}
\vspace{-0.1 cm}
Without loss of generality, we randomly select four domains from each of these two datasets, including "Treatment", "Inheritance", "New York City", and "Modern History", to construct different types of agents. 
Since the response from each target agent may not be completely matched with the correct response in the QA pair from the database, it is necessary to determine whether the response is a hallucination or not. We follow existing work~\cite{su-etal-2022-read} to use the GQA metric to identify the responses that include hallucinations. More specifically, a binary label is assigned according to the average of unigram F1 and ROUGE-L. A hallucination is labeled if this average is less than 0.5 \cite{10.1145/3583780.3614905}.

\vspace{-0.2 cm}
\subsection{Evaluation Setup} 


\begin{figure*}[htp]
    \centering
    \footnotesize
    \subfloat{
        \label{fig:abla1}
        \includegraphics[width=0.26\linewidth]{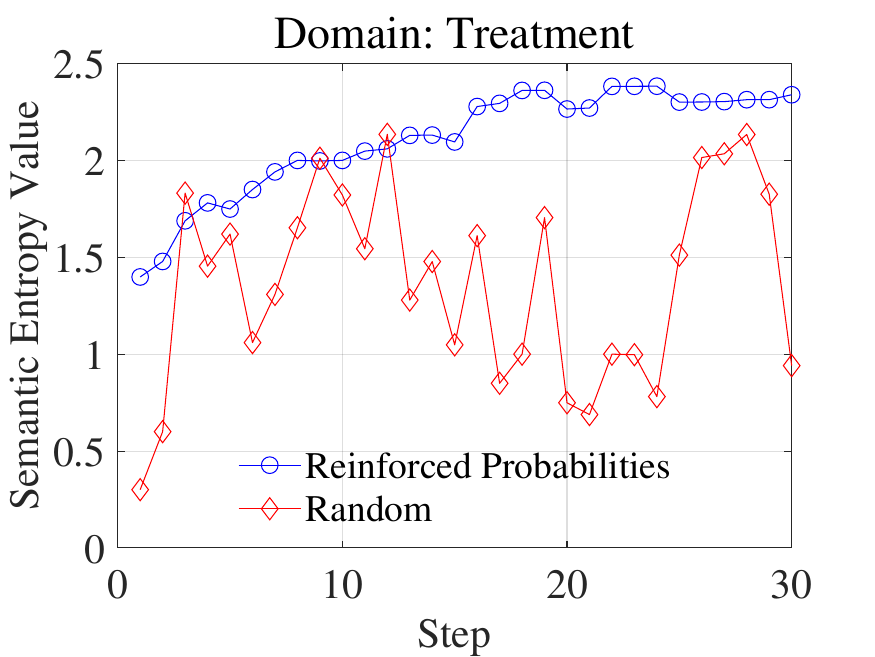}
    }
    \hspace{-0.6cm} 
    \subfloat{
        \label{fig:abla2}
        \includegraphics[width=0.26\linewidth]{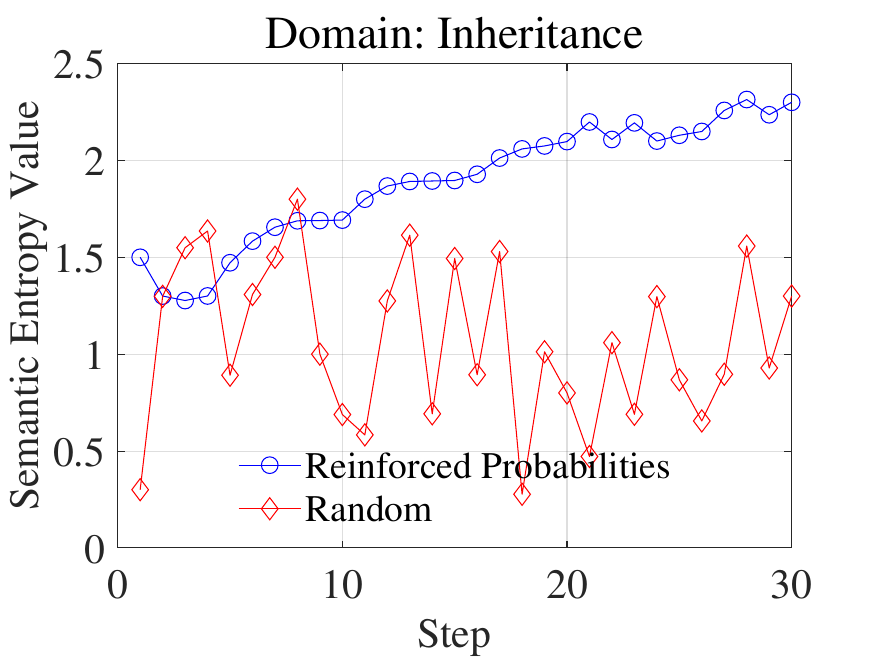}
    }
    \hspace{-0.6cm} 
    \subfloat{
        \label{fig:abla3}
        \includegraphics[width=0.26\linewidth]{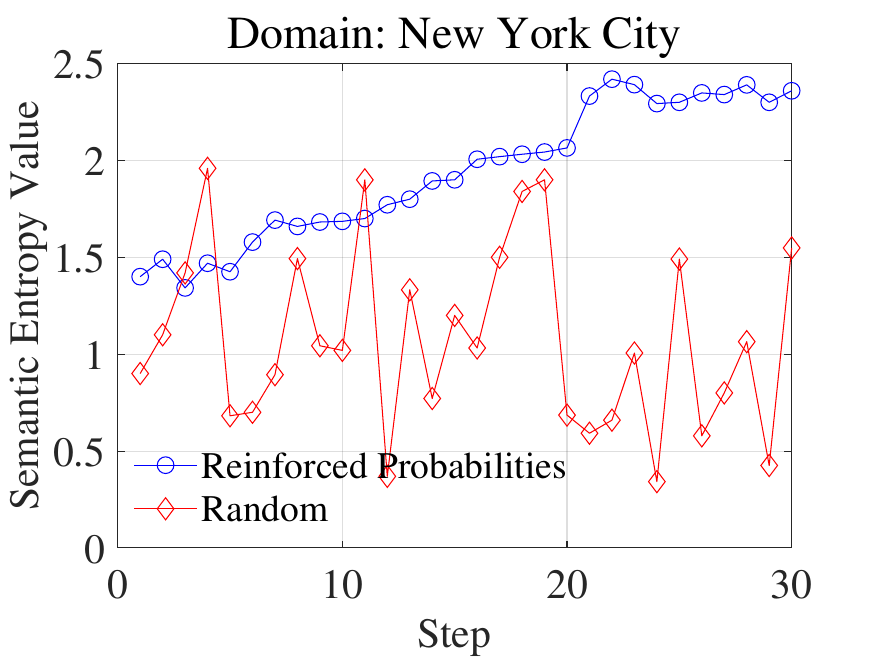}
    }
    \hspace{-0.6cm} 
    \subfloat{
        \label{fig:abla4}
        \includegraphics[width=0.26\linewidth]{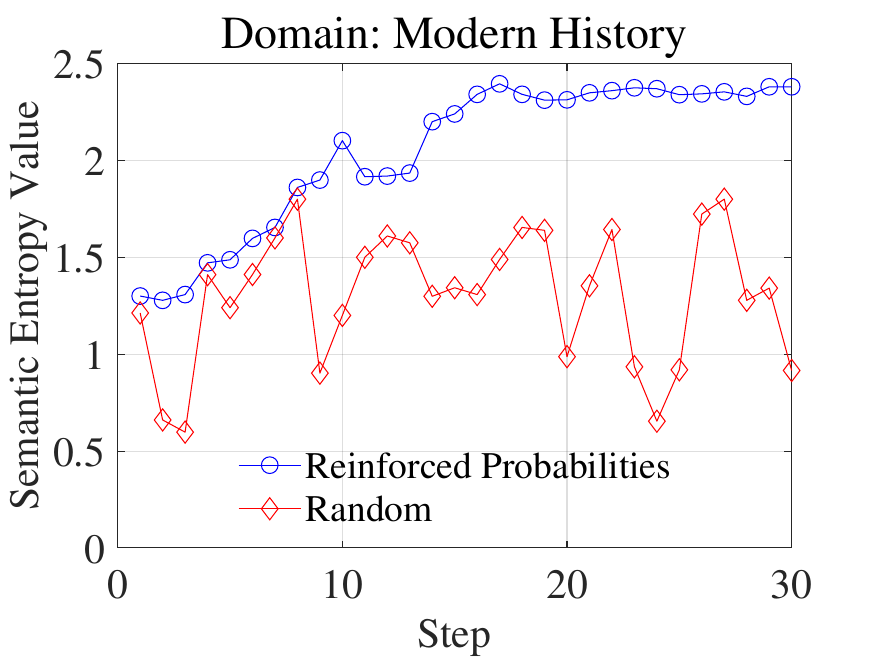}
    }
    \hspace{-0.6cm} 

    \caption{\label{fig:RL abla}Convergence of our exploration process.}

\end{figure*}


To evaluate the performance of $\ours$, six LLMs, including Llama (Llama2-7B-Instruct and Llama3.1-8B-Instruct), Mistral-7b, qwen2-1.5b, Falcon-7b and Vicuna-7b, are used to support agent inference. 
All agents in specific domains are implemented using RAG technology.
Specifically, this RAG pipeline is constructed with Elasticsearch \cite{elasticsearch} served as the vector database, which is used to record generalized bounds in the vector format to serve as references for identifying conditions where hallucinations occur. In addition, the m3e-base \cite{M3E} is used as the embedding model to vectorize the information of a generalized bound, including the query and the corresponding responses stored in the database that represent the bounds, as well as the input query in the hallucination monitoring process. We also utilize a repository within Agentscope V0.1.0 \cite{agentscope} to enable exploration of the generalization bound and monitoring of hallucinations of LLM-empowered agents.

During modeling the generalization bound, Qwen-max is used to generate queries, while GPT 4 is used to judge whether each response of the target LLM has hallucinations. In addition, we also incorporate a supervised method. When using GPT 4 for assistance in judgment, we evaluate the confidence in the inference results. If confidence is less than 60\%, we perform a manual review to ensure the accuracy of the judgment before proceeding \cite{casper2023open}.
During training the policy network of reinforcement determination on fractal probability, the learning rate is set to $10^{-4}$ and the batch size is 64. It converges in 300 epochs. 
We set the ratio $\gamma$ to 0.6 to guide the search for bounds, randomly initialize ten queries of the bound searching for each domain, and the similarity threshold $\epsilon$ to 0.8 for the monitoring of hallucinations.
The impacts of setting the two parameters to different values are studied in Section \ref{sec:ablation}.

Three popular hallucination detection methods are included as baselines: 1) Predictive Probability (PP)~\cite{manakul2023selfcheckgpt}, 2) In-Context-Learning Prompt (ICL)~\cite{xu2024misconfidence}, and 3) SelfCheckGPT (SCG)~\cite{manakul2023selfcheckgpt}. In addition, a comprehensive set of metrics is used in our evaluation, including: 1) the area under the receiver operator characteristic curve (AUROC), 2) the area under the precision recall curve (AUC-PR), 3) the F1 score, and 4) the accuracy. In addition, we also recorded the semantic entropy defined in Formula (2) in the Appendix of the output of the target agent to illustrate the uncertainty metric.

\vspace{-0.2 cm}
\subsection{Convergence Study of The Exploration}
\label{sec:Convergence}

Our probabilistic fractal exploration method is reinforced to explore towards the generalization bound. Through a reward mechanism based on increases in semantic entropy, our fractal exploration process is directed toward generating statements with higher uncertainty—indicating proximity to the generalization bound.  To evaluate the completeness of fractal exploration, we compare the performance of using the reinforced determination of fractal probabilities  against a baseline that uses randomly assigned probabilities. 

As shown in Figure \ref{fig:RL abla}, we present the semantic entropy values over the final 30 steps of the exploration process.  This can also be observed to investigate how semantic entropy evolves with the exploration of the generalization bound through fractal transformations. The results demonstrate that with reinforced fractal probability selection, each exploration step consistently increases or maintains semantic entropy, signaling effective converge on the generalization bound. 
In contrast, the random probability strategy yields volatile entropy values, with no clear trend, indicating an unreliable and less directed search process. These findings suggest that reinforcement-guided fractal exploration offers a more robust and targeted approach to identifying generalization boundaries.

\subsection{Effectiveness of Hallucination Monitoring}
We evaluated the effectiveness of $\ours$, and the results are presented in Table~\ref{tab1}. $\ours$ achieves the best performance in Inheritance and Modern History domains, demonstrating both its superiority and adaptability to various types of agent. Specifically, our method improves the AUROC and AUC-PR metrics up to 8\% over the best baseline, highlighting its effectiveness in distinguishing qualified output from hallucinations.
The only exception is the New York City topic, where SelfCheckGPT outperforms our method in monitoring hallucinations. This may be due to the miscellaneous slangy dialogues on this topic, which SelfCheckGPT appears to be better equipped to handle.
\begin{table*}[htpb]
\centering
\footnotesize
\caption{\label{tab1}Hallucination monitoring performance on different agents}
\begin{tabular}{lcccccccccc}
\hline
\multicolumn{1}{l|}{\textbf{}} & \multicolumn{1}{l|}{\textbf{Backbone}}                                                             & \multicolumn{1}{l}{\textbf{AUROC}$\uparrow$} & \multicolumn{1}{l}{\textbf{AUC-PR}$\uparrow$} & \multicolumn{1}{l}{\textbf{F1}$\uparrow$} & \multicolumn{1}{l|}{\textbf{Acc}$\uparrow$}  & \multicolumn{1}{l|}{\textbf{Backbone}}                                                               & \multicolumn{1}{l}{\textbf{AUROC}$\uparrow$} & \multicolumn{1}{l}{\textbf{AUC-PR}$\uparrow$} & \multicolumn{1}{l}{\textbf{F1}$\uparrow$} & \multicolumn{1}{l}{\textbf{Acc}$\uparrow$} \\ \hline
\multicolumn{11}{c}{Treatment}                                                                                                                              \\ \hline
\multicolumn{1}{l|}{PP}              & \multicolumn{1}{c|}{\multirow{4}{*}{\begin{tabular}[c]{@{}c@{}}Llama2\end{tabular}}} & 0.56                               & 0.54                                & 0.56                            & \multicolumn{1}{c|}{0.66}          & \multicolumn{1}{c|}{\multirow{4}{*}{\begin{tabular}[c]{@{}c@{}}Llama3.1\end{tabular}}} & 0.56                               & 0.69                                & 0.6                             & 0.65                             \\
\multicolumn{1}{l|}{ICL}             & \multicolumn{1}{c|}{}                                                                              & 0.59                               & 0.56                                & 0.60                            & \multicolumn{1}{c|}{0.55}          & \multicolumn{1}{c|}{}                                                                                & 0.48                               & 0.71                                & 0.7                             & 0.61                             \\
\multicolumn{1}{l|}{SCG}             & \multicolumn{1}{c|}{}                                                                              & 0.71                               & 0.72                                & 0.69                            & \multicolumn{1}{c|}{0.7}           & \multicolumn{1}{c|}{}                                                                                & 0.74                               & 0.84                                & 0.7                             & 0.75                             \\
\multicolumn{1}{l|}{\textbf{$\ours$}}   & \multicolumn{1}{c|}{}                                                                              & \textbf{\rev{0.76}}                      & \textbf{\rev{0.80}}                       & \textbf{\rev{0.79}}                   & \multicolumn{1}{c|}{\textbf{\rev{0.73}}} & \multicolumn{1}{c|}{}                                                                                & \textbf{\rev{0.80}}                      & \textbf{\rev{0.86}}                       & \textbf{\rev{0.82}}                   & \textbf{\rev{0.88}}                    \\ \hline
\multicolumn{11}{c}{Inheritance}                                                                                                                                   \\ \hline
\multicolumn{1}{l|}{PP}              & \multicolumn{1}{c|}{\multirow{4}{*}{\begin{tabular}[c]{@{}c@{}}Llama2\end{tabular}}} & 0.68                               & 0.60                                & 0.77                            & \multicolumn{1}{c|}{0.75}          & \multicolumn{1}{c|}{\multirow{4}{*}{\begin{tabular}[c]{@{}c@{}}Llama3.1\end{tabular}}} & 0.71                               & 0.79                                & 0.72                            & 0.71                             \\
\multicolumn{1}{l|}{ICL}             & \multicolumn{1}{c|}{}                                                                              & 0.69                               & 0.67                                & 0.59                            & \multicolumn{1}{c|}{0.56}          & \multicolumn{1}{c|}{}                                                                                & 0.61                               & 0.73                                & 0.65                            & 0.65                             \\
\multicolumn{1}{l|}{SCG}             & \multicolumn{1}{c|}{}                                                                              & 0.70                               & 0.73                                & 0.68                            & \multicolumn{1}{c|}{0.75}          & \multicolumn{1}{c|}{}                                                                                & 0.85                               & 0.84                                & \textbf{\rev{0.85}}                   & 0.85                             \\
\multicolumn{1}{l|}{\textbf{$\ours$}}   & \multicolumn{1}{c|}{}                                                                              & \textbf{\rev{0.70}}                      & \textbf{\rev{0.79}}                       & \textbf{\rev{0.8}}                    & \multicolumn{1}{c|}{\textbf{\rev{0.78}}} & \multicolumn{1}{c|}{}                                                                                & \textbf{\rev{0.90}}                      & \textbf{\rev{0.86}}                       & 0.82                            & \textbf{\rev{0.88}}                    \\ \hline
\multicolumn{11}{c}{New York City}                                                  \\ \hline
\multicolumn{1}{l|}{PP}              & \multicolumn{1}{c|}{\multirow{4}{*}{\begin{tabular}[c]{@{}c@{}}Llama2\end{tabular}}} & 0.54                               & 0.12                                & 0.21                            & \multicolumn{1}{c|}{0.74}          & \multicolumn{1}{c|}{\multirow{4}{*}{\begin{tabular}[c]{@{}c@{}}Llama3.1\end{tabular}}} & 0.60                               & 0.58                                & 0.52                            & 0.53                             \\
\multicolumn{1}{l|}{ICL}             & \multicolumn{1}{c|}{}                                                                              & 0.59                               & 0.22                                & 0.32                            & \multicolumn{1}{c|}{0.75}          & \multicolumn{1}{c|}{}                                                                                & 0.58                               & 0.64                                & 0.63                            & 0.45                             \\
\multicolumn{1}{l|}{SCG}             & \multicolumn{1}{c|}{}                                                                              & 0.82                               & 0.72                                & \textbf{\rev{0.85}}                   & \multicolumn{1}{c|}{0.82}          & \multicolumn{1}{c|}{}                                                                                & 0.86                               & \textbf{\rev{0.84}}                       & 0.87                            & \textbf{\rev{0.86}}                    \\
\multicolumn{1}{l|}{\textbf{$\ours$}}   & \multicolumn{1}{c|}{}                                                                              & \textbf{\rev{0.88}}                      & \textbf{\rev{0.77}}                       & 0.75                            & \multicolumn{1}{c|}{\textbf{\rev{0.89}}} & \multicolumn{1}{c|}{}                                                                                & \textbf{\rev{0.89}}                      & 0.82                                & \textbf{\rev{0.88}}                   & 0.84                             \\ \hline
\multicolumn{11}{c}{Modern History}                                                                                                                               \\ \hline
\multicolumn{1}{l|}{PP}              & \multicolumn{1}{c|}{\multirow{4}{*}{\begin{tabular}[c]{@{}c@{}}Llama2\end{tabular}}} & 0.74                               & 0.76                                & 0.73                            & \multicolumn{1}{c|}{0.79}          & \multicolumn{1}{c|}{\multirow{4}{*}{\begin{tabular}[c]{@{}c@{}}Llama3.1\end{tabular}}} & 0.48                               & 0.45                                & 0.46                            & 0.56                             \\
\multicolumn{1}{l|}{ICL}             & \multicolumn{1}{c|}{}                                                                              & 0.69                               & 0.67                                & 0.64                            & \multicolumn{1}{c|}{0.67}          & \multicolumn{1}{c|}{}                                                                                & 0.59                               & 0.55                                & 0.54                            & 0.61                             \\
\multicolumn{1}{l|}{SCG}             & \multicolumn{1}{c|}{}                                                                              & 0.78                               & 0.79                                & \textbf{\rev{0.81}}                   & \multicolumn{1}{c|}{0.82}          & \multicolumn{1}{c|}{}                                                                                & 0.78                               & 0.77                                & 0.6                             & 0.76                             \\
\multicolumn{1}{l|}{\textbf{$\ours$}}   & \multicolumn{1}{c|}{}                                                                              & \textbf{\rev{0.84}}                      & \textbf{\rev{0.80}}                       & 0.77                            & \multicolumn{1}{c|}{\textbf{\rev{0.89}}} & \multicolumn{1}{c|}{}                                                                                & \textbf{\rev{0.84}}                      & \textbf{\rev{0.84}}                       & \textbf{\rev{0.67}}                   & \textbf{\rev{0.89}}                    \\ \hline
\end{tabular}
\end{table*}

Compared to baselines that use a fixed threshold for hallucination detecting, $\ours$ shows greater effectiveness across agent hallucinations in different domains. The Treatment and Inheritance domains primarily involve scientific knowledge, while New York City and Modern History consist of miscellaneous questions. $\ours$ performs particularly well in domains that allow for divergent responses, probably because it better adapts to the semantic diversity and complexity inherent in such queries.
Finally, as a basic scheme only relying on LLM to detect hallucinations, ICL performs poorly for hallucination identification, while the used LLMs struggle to reliably detect their own hallucinations without external guidance.

To evaluate the effectiveness of $\ours$ in monitoring hallucinations among agents empowered by other LLM models, Table~\ref{tab2} presents the hallucination monitoring performance for agents with the other four models, Mistral-7b, Qwen2-1.5b, Falcon-7b and Vicuna-7b. Without loss of generalization, the agents in Treatment domain are selected to construct experiments. The experimental results demonstrate that $\ours$ consistently outperforms the baselines, achieving the highest accuracy and F1 scores. In particular, $\ours$ shows the most significant improvement for agents empowered by Qwen2-1.5b, reaching the highest accuracy of 0.85. 
PP and ICL show lower monitoring accuracy in agents with Vicuna, SelfCheckGPT, and $\ours$ experiences a marked increase in F1 score. These results demonstrate the superiority of $\ours$ as the most robust approach, enhancing its performance across a variety of LLM architectures.

\begin{table}[htbp]

\caption{\label{tab2}Additional results for evaluating hallucination monitoring performance}
\begin{tabular}{llllll}
\hline
Model                       & Method & AUROC  $\uparrow$       & AUC-PR $\uparrow$       & Acc  $\uparrow$         & F1   $\uparrow$         \\ \hline
\multirow{4}{*}{Mistral-7b} & PP     & 0.43          & 0.49          & 0.56          & 0.37          \\
                            & ICL    & 0.58          & 0.52          & 0.59          & 0.49          \\
                            & SCG    & 0.56          & 0.67          & 0.69          & 0.64          \\
                            & \textbf{$\ours$}   & \textbf{\rev{0.69}} & \textbf{\rev{0.70}} & \textbf{\rev{0.79}} & \textbf{\rev{0.77}} \\ \hline
\multirow{4}{*}{qwen2-1.5b} & PP     & 0.49          & 0.50          & 0.50          & 0.42          \\
                            & ICL    & 0.52          & 0.39          & 0.54          & 0.44          \\
                            & SCG    & 0.72          & 0.80          & 0.78          & 0.68          \\
                            & \textbf{$\ours$}   & \textbf{\rev{0.79}} & \textbf{\rev{0.80}} & \textbf{\rev{0.85}} & \textbf{\rev{0.81}} \\ \hline
\multirow{4}{*}{Falcon-7b}  & PP     & 0.64          & 0.59          & 0.49          & 0.44          \\
                            & ICL    & 0.57          & 0.60          & 0.51          & 0.55          \\
                            & SCG    & 0.68          & 0.71          & 0.75          & 0.79          \\
                            & \textbf{$\ours$}   & \textbf{\rev{0.73}} & \textbf{\rev{0.75}} & \textbf{\rev{0.79}} & \textbf{\rev{0.80}} \\ \hline
\multirow{4}{*}{Vicuna-7b}  & PP     & 0.39          & 0.41          & 0.45          & 0.67          \\
                            & ICL    & 0.60          & 0.58          & 0.56          & 0.79          \\
                            & SCG    & 0.68          & \textbf{\rev{0.72}} & 0.71          & 0.81          \\
                            & \textbf{$\ours$}   & \textbf{\rev{0.75}} & 0.71          & \textbf{\rev{0.84}} & \textbf{\rev{0.89}} \\ \hline
\end{tabular}
\end{table}

\vspace{-0.2 cm}
\subsection{Ablation Study}
\label{sec:ablation}

\begin{figure}[H]
\vspace{-0.3 cm}
    \centering
    \subfloat[]{
        \label{fig:llama2}
        \includegraphics[width=0.34\linewidth]{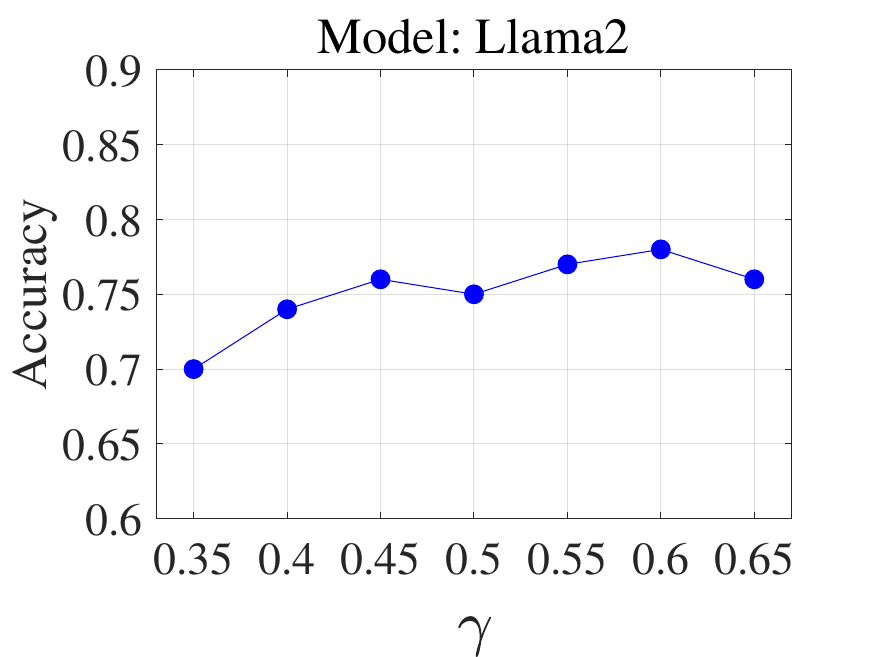}
    }
    \hspace{-0.5 cm}
    \subfloat[]{
        \label{llama3}
        \includegraphics[width=0.34\linewidth]{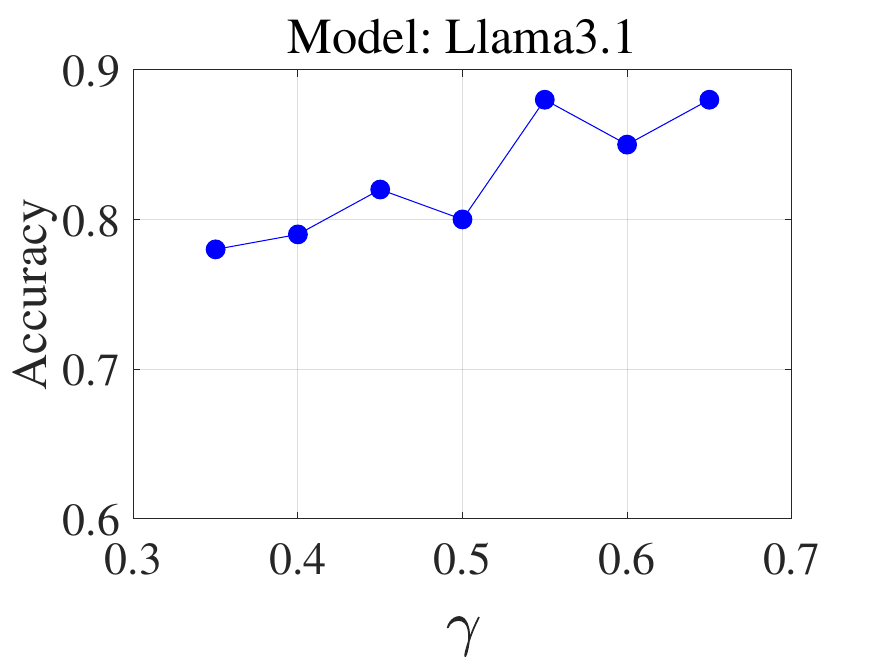}
    }
    \hspace{-0.5 cm}
    \subfloat[]{
        \label{epsilon}
        \includegraphics[width=0.34\linewidth]{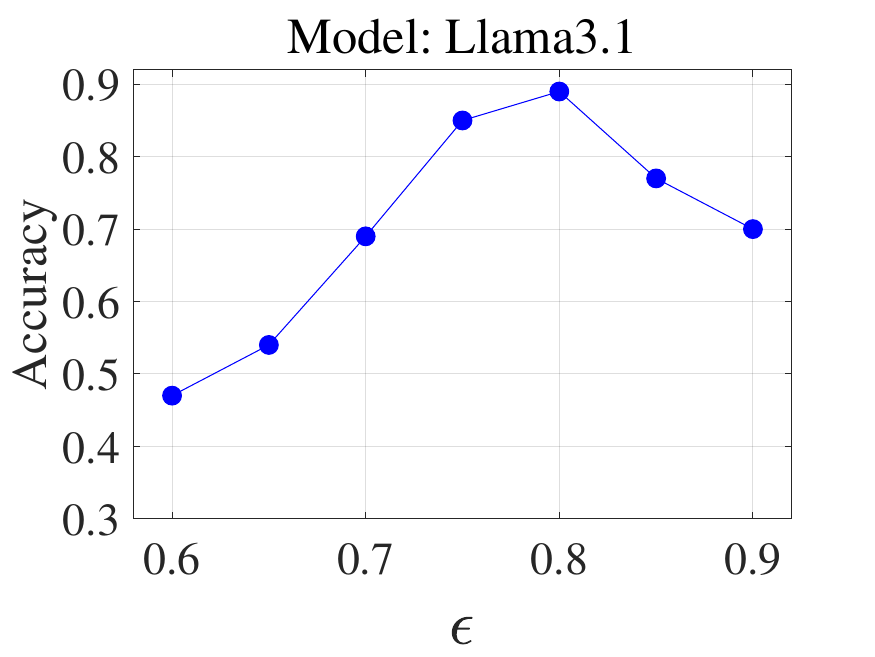}
    }

    \caption{\label{fig:Ablation} Ablation Study of $\gamma$ and $\epsilon
    $.}

\vspace{-0.2 cm}
\end{figure}

We investigate the impact of parameter $\gamma$ and $\epsilon$ 
on the monitoring accuracy in the Inheritance domain. As shown in Figures \ref{fig:llama2} and \ref{llama3}, when $\gamma$ varies from 0.35 to 0.65, the monitoring accuracy remains consistently high, maintaining a level between 0.78 and 0.88. The curve exhibits remarkable stability across different $\gamma$ values, with only a slight fluctuation. The stable performance across a wide range of $\gamma$ values demonstrates that our proposed monitoring method is relatively insensitive to the choice of $\gamma$. This robustness is particularly valuable for practical applications, as it suggests that the method can maintain reliable performance without requiring precise fine-tuning of $\gamma$. 
Similarly, the impact of the parameter $\epsilon$ on the accuracy of the monitoring is evaluated, and the tested values of $\epsilon$ vary between 0.6 and 0.9. As illustrated in Figure \ref{epsilon}, $\epsilon$ increases from 0.6 to 0.8, the accuracy improves, and the highest performance is achieved at $\epsilon$ of $0.8$. These observations suggest that $\epsilon = 0.8$ strikes an optimal trade-off, yielding the best performance. 

\section{Related Work}
\vspace{-0.1 cm}
\label{sec:relate}


Most of the related approaches take LLM as a white box to detect hallucinations.
Ji et al. \cite{ji2024llm} used a mutual information-based feature selection method to select sensitive neurons from the last activation layer and trained a classifier using Llama MLP. Han et al. \cite{han2024semantic} proposed a method to approximate semantic entropy from the hidden state of the model, converting the semantic entropy into binary labels, and trained a logistic regression classifier to predict hallucinations. Zhu et al. \cite{zhu2024pollmgraph} used PCA to reduce the dimensionality of the hidden layer embedding and adopted interval partitioning or GMM clustering to establish abstract states. In addition, they also used Markov models and hidden Markov models to capture state transitions, and used a small amount of annotated reference data to link internal state transitions with hallucination/factual output behaviors. He et al. \cite{he-etal-2024-llm} combined the static features within the model with the dynamic features and used Siamese networks to identify situations where the answers of the large language model deviate from the facts. 
Gaurang et al. \cite{sriramanan2024llm} proposes a hallucination detection method that analyze internal model signals. Xiaoling et al. \cite{zhouhademif} propose HADEMIF for detecting hallucinations in LLMs by leveraging a Deep Dynamic Decision Tree and an MLP to calibrate model predictions. 
Requiring the access to internal states of LLMs to detect hallucinations, these methods not only suffer from high complexity and computational demand but also may not be feasible for commercial LLM software. This highlights the need for research on the detection of hallucinations without accessing the internal states of the LLMs.

The other solutions detect hallucinations whereas considering the black-box nature of LLMs.
Hou et al. \cite{hou2024probabilistic} proposed using the belief tree, a probabilistic framework, to detect hallucinations using the logical consistency between model beliefs. Quevedo et al. \cite{quevedo2024detecting} extracted features using two LLMs and used these features to train logistic regression and simple neural networks to detect hallucinations. 
Although these methods detect hallucination through output features or associated confidence scores, the limited understanding of the generalization bound of LLMs leads to poorly calibrated confidence estimates. 

\vspace{-0.2 cm}
\section{Conclusion}
\label{sec:conclusion}
\vspace{-0.1 cm}
In this work, we present an in-depth study of the hallucination phenomenon in LLM-empowered agents through the lens of fine-grained domains. Based on our findings, we propose an effective and efficient hallucination monitor $\ours$, according to a few key observations from our study. Our research reveals that LLMs exhibit similar generalization bounds within the same domain, providing a foundation for accurately monitoring hallucinations in specific domains. To take advantage of this key insight, we have designed a reinforced probabilistic fractal exploration method that efficiently identifies the generalization bound of an LLM-empowered agent within a domain. Despite the black-box nature of LLM-empowered agents, this approach significantly accelerates the boundary identification process while improving both the accuracy and efficiency of hallucination monitoring based on the generalization bound.
Extensive experimental evaluations demonstrate that our method outperforms existing mainstream hallucination monitoring techniques across multi-topic datasets and different foundation LLMs. This work not only offers a novel technical pathway for monitoring hallucinations for agents in a specific domain, but also provides robust theoretical and practical support to enhance their security and dependability in critical applications. 





\bibliography{mybibfile}


\end{document}